\documentclass[letterpaper, 10 pt, conference]{ieeeconf}  

\usepackage{blindtext}
\usepackage{graphicx}
\usepackage{times}
\usepackage{helvet}
\usepackage{courier}
\usepackage{mathrsfs}
\usepackage{subfigure}
\usepackage{amsmath}
\usepackage{amssymb}
\usepackage{enumerate}
\usepackage{epsfig}
\usepackage{cite}
\usepackage{epstopdf}
\usepackage{algorithm}
\usepackage{algorithmic}
\usepackage{clrscode}
\usepackage{pifont}
\usepackage{amssymb}
\usepackage{diagbox}
\usepackage{multirow}
\usepackage{multicol}
\IEEEoverridecommandlockouts                              

\newtheorem{ass}{\textbf{Assumption}}

\newtheorem{dnt}{\textbf{Definition}}
\newtheorem{rem}{\textbf{Remark}}

\usepackage{xcolor}

\usepackage{lipsum} 
\graphicspath{{figures/}} 
\usepackage{footmisc}
\usepackage{url}

\title{\LARGE \bf
Constrained Passive Interaction Control: Leveraging \\ Passivity and Safety for Robot Manipulators}

\author{Zhiquan Zhang$^{\dagger}$, Tianyu Li$^{\dagger}$ and Nadia Figueroa$^{\dagger}$
\thanks{$^{\dagger}$Zhiquan Zhang, Tianyu Li and Nadia Figueroa are with School of Engineering and Applied Science, University of Pennsylvania, Pennsylvania, PA 19104 USA (email: \{zzq2000, tianyuli, nadiafig\}@seas.upenn.edu)}
}

\begin{document}

\maketitle
\thispagestyle{empty}
\pagestyle{empty}

\begin{abstract}
Passivity is necessary for robots to fluidly collaborate and interact with humans physically. Nevertheless, due to the unconstrained nature of passivity-based impedance control laws, the robot is vulnerable to infeasible and unsafe configurations upon physical perturbations. In this paper, we propose a novel control architecture that allows a torque-controlled robot to guarantee safety constraints such as kinematic limits, self-collisions, external collisions and singularities and is passive only when feasible. This is achieved by constraining a dynamical system based impedance control law with a relaxed hierarchical control barrier function quadratic program subject to multiple concurrent, possibly contradicting, constraints. Joint space constraints are formulated from efficient data-driven self- and external $\mathcal{C}^2$ collision boundary functions. We theoretically prove constraint satisfaction and show that the robot is passive when feasible. Our approach is validated in simulation and real robot experiments on a 7DoF Franka Research 3 manipulator.
\end{abstract}

\section{INTRODUCTION}
Robots that will physically interact with humans and execute tasks in dynamic human spaces, must allow any type of physical perturbations imposed on them  by the humans, and seamlessly recover from them to achieve the intended (or changed) task. This requires robots to not only be inherently compliant but also passive to such external perturbations. The concept of passivity was first introduced in the context of robotics by Hogan and Colgate \cite{Hogan1984,Colgate1988} to analyze stability issues related to impedance controllers. When a robot is physically interacting with an unknown environment or a human with unknown intention or dynamics the standard Lyapunov-style tools for stability analysis are difficult to apply. Analyzing passivity becomes a more intuitive and indirect way of ensuring stability of physically interactive systems \cite{vanderSchaft2017}. 
Thus, a controller designed to preserve a passive relation between external forces and robot velocity ensures stability both in free motion and in contact with passive environments \cite{Colgate1988}. A more general notion of passivity is the property of a dynamical system to not produce more energy than it receives \cite{spong2nd}. Hence, the majority of passivity-based controllers leverage energy exchange techniques \cite{Energy2017} built on impedance-based control laws \cite{Hogan1984,Stramigioli1999,Li1999,Kishi2003,Duindam2004,AlbuSchaffer2007,Ferraguti2013,kronander2015passive,Ferraguti2015,AbuDakka2020}. Of particular interest is ensuring passivity/stability for variable impedance control (VIC) which is often used in physical human-robot interaction and teleoperation in tasks that require adaptation of the robot's stiffness \cite{AbuDakka2020}. Such change in stiffness, when unbounded may induce energy-injecting terms which can destabilize a robot \cite{Ferraguti2013}. Several works have proposed techniques to preserve passivity in VIC, either by using tank-based \cite{Ferraguti2013} or port-Hamiltonian representations \cite{Ferraguti2015}, or deriving restrictions on damping and stiffness fluctuations via a modified Lyapunov function \cite{7560657}. 

Nevertheless, classic impedance control schemes that are driven by time-indexed reference trajectories can only be passive in the regulation case, loosing passivity during tracking -- a critical drawback of classic impedance control. A solution to this is to encode tasks as time-independent velocity fields, as initially advocated in \cite{Li1999,Kishi2003,Duindam2004}. The modern solution to this problem is to follow the dynamical system (DS) based motion planning approach where the desired task is encoded as a first order autonomous DS \cite{dsbook}. Then a negative velocity feedback control law is used to drive the robot along the directions of the DS while dissipating energy in orthogonal directions, ensuring passive interaction control via a storage function that includes the potential function used to shape the DS \cite{kronander2015passive}. While this approach has the properties we desire, it has one major drawback -- it cannot ensure the safety of the robot upon physical perturbations, such as violating joint limits, self-collisions or collisions with external objects. Hence, the robot is vulnerable to violating all of these constraints which might lead to motor failures, undesirable collisions or even injuries to other humans. In this work, we posit that a robot operating in a human-centric environment should be passive only when feasible and  when none of these constraints are being violated such that safe operation of the robot is always ensured. Such a proposition could be thought of as a realization of Asimov's third law. 

\begin{figure}[!tbp]
  \centering
  \includegraphics[width=\linewidth]{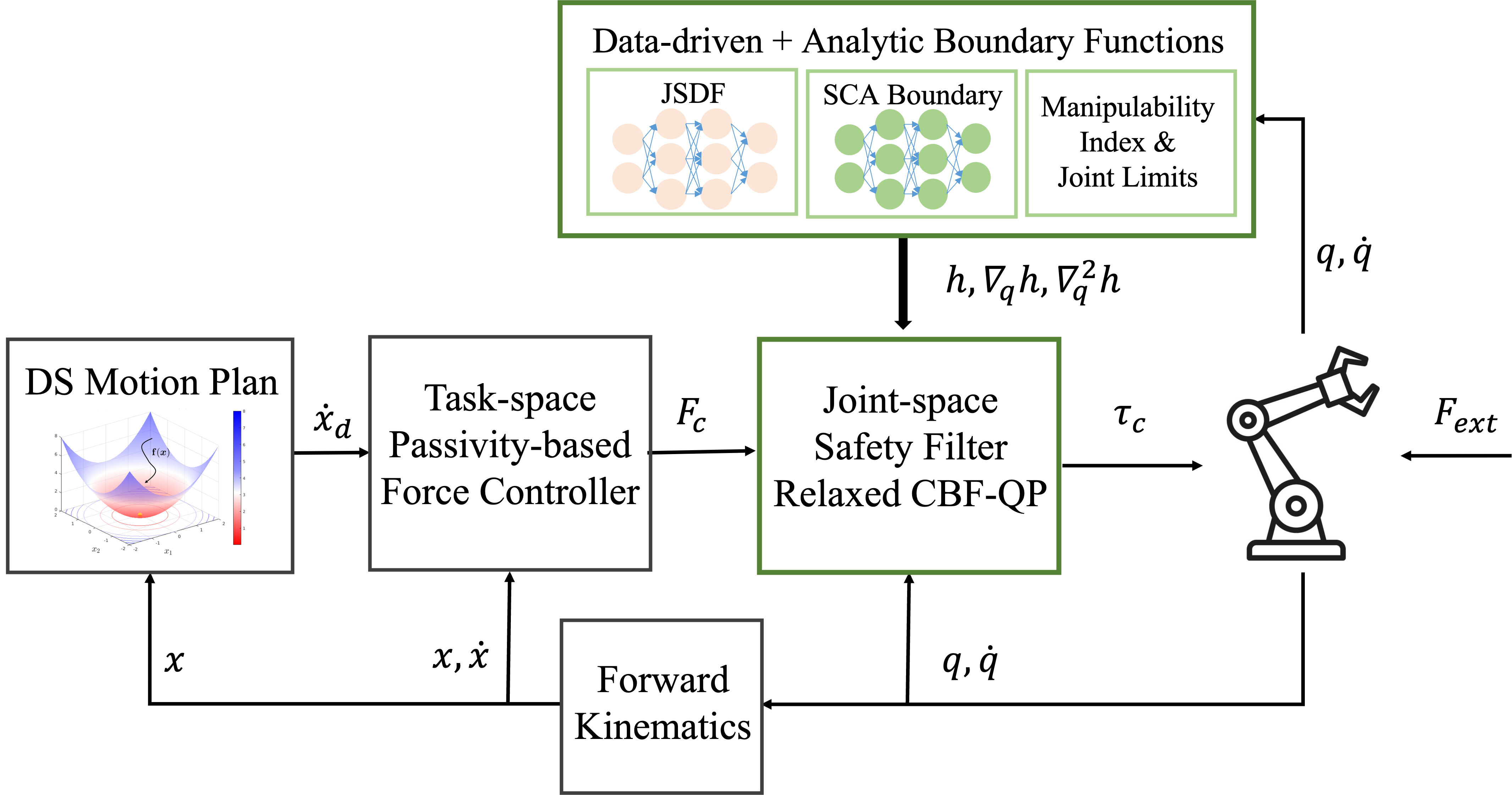}
    \caption{Schematic of our proposed constrained passive interaction control framework. Green blocks denote novel components. Note that every block has a feedback term, from joint-space to task-space layers.}
  \vspace{-17pt}
\end{figure}

Note that this passive impedance control law \cite{kronander2015passive}, and in general most passivity-based controllers, operate at the force/torque level as they are designed for compliant torque-controlled robots. One can add repulsive torques to the passive impedance control laws to ensure collision avoidance \cite{khatib1986APF} or nullspace torques favor joint limit or singularity avoidance \cite{Dietrich2015NullTorque} however these approaches will either break passivity and cannot guarantee the satisfaction of constraints.

In recent years, control barrier functions \cite{ames2019cbf} have risen as the prominent tool to verify and enforce safety properties in the context of optimization-based safety-critical controllers for autonomous systems. Optimization-based CBF controllers to ensure safe control of robot manipulators when tracking a desired reference trajectory have been proposed \cite{7759067,9636794}, yet they are not passive, only operate in the task space of the robot and do not ensure joint-space constraints. 

\textbf{Contributions} In this work, we propose an optimization-based CBF control framework that allows a robot to remain passive to external perturbations following the DS-based passive interaction controller \cite{kronander2015passive}, yet is guaranteed to satisfy joint-space constraints such as, joint limits, self-collisions, external collisions and singularities. To formulate joint-space collision avoidance constraints we adopt the data-driven approaches proposed in prior work \cite{mirrazavi2018sca-dual,koptev2021sca-humanoid,dsbook,koptev2023njsdf} and learn $C^2$ differentiable self- and external collision boundary functions by sampling the robot's joint-space. Such boundary functions have only been used in motion-level controllers such as IK and MPC. The only work closest to ours is \cite{Kurtz2021} which proposes a similar optimization-based approach that guarantees passivity and single singularity avoidance constraint. In our framework, passivity is formulated as a task-space objective which is guaranteed only when feasible, while joint-space constraints are always satisfied and prioritized. 


\section{PROBLEM STATEMENT}\label{section2}
We seek to control an $n$-DoF manipulator with rigid-body dynamics derived by the Euler-Lagrange equations \cite{spong2nd}:
\begin{equation}\label{eq1}
	H(q)\ddot{q} + C(q, \dot{q})\dot{q} + G(q) = \tau_c + \tau_{ext},
\end{equation}
where $q,\dot{q}, \ddot{q} \in \mathbb{R}^n$ denote the joint positions, velocities, and accelerations. $H(q)\in \mathbb{R}^{n\times n}$, $C(q, \dot{q})\in \mathbb{R}^{n\times n}$, $G(q)\in \mathbb{R}^{n}$ denote the inertia matrix, Coriolis matrix and gravity respectively. $\tau_c\in \mathbb{R}^n$ and $\tau_{ext}\in \mathbb{R}^n$ denote the control input and external torque applied on the joints. 

Given a desired $d$-dimensional\footnote{$d=3$ for end-effector position and $d=6$ for position and orientation} task-space control input $F_c\in\mathbb{R}^d$ one can compute the desired control torques as $\tau_c=J(q)^{\top}F_c$ where $J(q)\in \mathbb{R}^{d\times n}$ is the manipulator Jacobian. Similarly, external task-space forces are mapped to torques as follows $\tau_{ext}=J(q)^{\top}F_{ext}$ and joint-space velocities mapped to task-space as $\dot{x}=J(q)\dot{q}$ with $x,\dot{x}\in\mathbb{R}^d$ being the task-space state and velocity of the manipulator.

The \textbf{goal} of this work is to design a controller $\tau_c$ for \eqref{eq1} with the following \textbf{task-space objective}:
\begin{itemize}
\item[$\mathcal{O}_1$] \textit{Tracking:} Track a task-space reference velocity $\dot{x}_d\in\mathbb{R}^d$
\item[$\mathcal{O}_2$] \textit{Passivity:} The controlled-system should be passive in the robot's task space $(F_{ext},\dot{x})$ when feasible.
\end{itemize}
\noindent while guaranteeing the following \textbf{joint-space constraints}:
\begin{itemize}
\item[$\mathcal{C}_1$] \textit{Joint Limits Avoidance:} The robot's joint positions should remain within kinematic limits $q^{-}\leq q \leq q^{+}$.
\item[$\mathcal{C}_2$] \textit{Self-Collision Avoidance:} The controller should avoid self-collided configurations; i.e., the joint configurations should lie in the self-collision free configuration set $\mathcal{Q}_{SCA} = \{q_{SCA} \in \mathbb{R}^n~|~\min(d_{ij}(q)) \ge 0 \}$, where $d_{ij}(q)$, $i = 1, \cdots, n$, $j = 1, \cdots, n$, $i \neq j$, is the minimal distance between link-$i$ and link-$j$.  
\item[$\mathcal{C}_3$] \textit{External Collision Avoidance:} The controller should avoid configurations where the robot collides with an external obstacle; i.e., the joint configurations should lie in the external collision free set $\mathcal{Q}_{ECA} = \{q_{ECA} \in \mathbb{R}^n~|~\min\{d_{ext}(q)\ge 0\}\}$, where $d_{ext}$ denotes the distance between the body of the manipulator and the possible collided environment in cartesian space.
\item[$\mathcal{C}_4$] \textit{Singularity Avoidance:} Avoid joint configurations that lead to singularities; i.e., remain within the singularity avoidance set $\mathcal{Q}_{SA} = \{q \in \mathbb{R}^n~|~\text{rank}(J(q))\geq d\}$.
\end{itemize}


\section{PRELIMINARIES}\label{section3}
Next we describe the control theory tools used in this work to achieve $\mathcal{O}_1-\mathcal{O}_2$ while satisfying constraints $\mathcal{C}_1-\mathcal{C}_4$.
\vspace{-5pt}
\subsection{Dynamical System Motion Planning} 
In this work, we adopt the dynamical system (DS) based motion planning approach \cite{dsbook}. The desired task-space motion of the robot is defined in $x\in \mathbb{R}^d$ and constrained by a system of ODEs. Let $f(x)$ be a first-order, autonomous DS describing the robot's nominal motion plan as, 
\begin{equation}
\label{eq2}
\dot{x} = f(x) ~~~\text{s.t.} \lim_{t\rightarrow\infty}|| x - x^*||= 0 
\end{equation}
with $f(x):\mathbb{R}^d\rightarrow\mathbb{R}^d$ being a continuously differentiable vector-valued function representing a DS that asymptotically converges to a single stable attractor (target) $x^*\in\mathbb{R}^d$. Hence, in this work the desired task-space velocity is $\dot{x}_d=f(x)$. 



\subsection{Passivity-Based Control with Dynamical Systems \cite{kronander2015passive}}
\label{section3b}
To ensure task-space passivity ($\mathcal{O}_2$) while tracking a desired DS motion plan ($\mathcal{O}_1$) we use the following velocity-based impedance control law proposed in \cite{kronander2015passive}
\begin{equation}\label{eq3}
	F_c = G_x(x) - D(x)(\dot x - f(x)),
\end{equation}
where $G_x(x) = J(q)^{-\top}G(q)$ is the gravity vector in task space and $D(x)\in\mathbb{R}^{d \times d}$ represents a task-space damping matrix. This negative velocity error feedback controller \eqref{eq2} makes the robot track $f(x)$ by generating kinetic energy along the desired direction of motion, and dissipates kinetic energy in directions perpendicular to it. This is achieved given that the following assumptions hold:

\begin{ass}
The DS $f(x)$ is conservative, i.e., shaped by the negative gradient of a potential function $\mathcal{P}\in \mathbb{R}^d \rightarrow \mathbb{R}$:
\vspace{-10pt}
\begin{equation}
\label{eq4}
   f(x) = -\nabla_x \mathcal{P}(x).
\end{equation}
\end{ass}
\begin{ass}
The damping matrix $D(x)\in\mathbb{R}^{d\times d}$ is a positive semi-definite matrix defined as, 
	\begin{equation}
 \label{eq5}
		D(x) = V(x)\Lambda V(x)^\top,
	\end{equation}
 where $\Lambda \in \mathbb{R}^{d\times d}$ is a diagonal matrix of non-negative values and $V(x) = [v_1, v_2, v_3] \in \mathbb{R}^{d\times d}$ is an orthonormal basis matrix constructed such that $v_1 = \frac{f(x)}{||f(x)||}$ follows the direction of the desired motion and the remaining vectors in the basis are mutually orthogonal and normalized.
\end{ass}


Using a storage function $\mathcal{S} = \frac{1}{2}\dot x^\top H_x \dot x + \lambda_1 \mathcal{P}(x)$ where $H_x= J^{-\top}H(q)J^{-1}$ is the task-space inertia matrix and $\lambda_1$ is the first diagonal element of $\Lambda$ from \eqref{eq5}, the robot described by system \eqref{eq1} under control \eqref{eq2} is proven to be passive wrt. input-output pair $(F_{ext},\dot{x})$
\cite{kronander2015passive}. Note that, Assumption 1 can be relaxed and passivity can still be proven as long as a conservative component can be extracted and by adding a stored energy variable to the storage function \cite{kronander2015passive}.

\vspace{-2.5pt}
\subsection{QP-based Controller via Control Barrier Functions \cite{ames2019cbf}}
In this work, we propose a convex optimization problem to solve for $\mathcal{O}_1-\mathcal{O}_2$ subject to constraints $\mathcal{C}_1-\mathcal{C}_4$ formulated as control barrier functions (CBF). Next we introduce the preliminaries on formulating such an optimization problem in terms of a general state variable $\xi \in \mathbb{R}^n$. Consider a general continuous-time nonlinear control affine system:
\begin{equation}
\label{eq6}
	\dot \xi = \underline{f}(\xi) + g(\xi)u,
\end{equation}
where $u\in \mathbb{R}^m$ denotes the control input. $\underline{f}(\xi)\in \mathbb{R}^n$ is the drift dynamics and $g(\xi)\in \mathbb{R}^{n\times m}$ is the input dynamics. Note that $\underline{f}(\xi)$ is not the same dynamics as the DS motion plan \eqref{eq2}. In this work, $\underline{f}(\xi)$ is derived from the robot dynamics \eqref{eq1}. We assume $\underline{f} + g(\xi)u$ is bounded and Lipschitz continuous. 

To enforce safety on \eqref{eq6}, we must define the invariant (safe) set $\mathcal{C}$, which is the superlevel set of a smooth function $h(\xi):\mathbb{R}^n\rightarrow\mathbb{R}$  with $\nabla_\xi h\neq0 ~\forall \xi$ such that $h(\xi)=0$ only at the set boundary $\partial \mathcal{C}$; i.e., $\mathcal{C} = \{\xi\in\mathbb{R}^n|h(\xi)\ge0\}$. $h(\xi)$ is, thus, a function that measures the distance from the controlled system state $\xi$ to the constraint boundary $\partial \mathcal{C}$. Nagumo proposed the following necessary and sufficient condition for set invariance \cite{ames2019cbf},
\begin{dnt}[Set Invariance]
\begin{equation}
\mathcal{C} ~~\text{is set invariant} \iff \Dot{h}(\xi) \geq 0 \; \forall \xi \in \partial \mathcal{C} \label{eq7}
\end{equation}
\end{dnt}

One must then generate an admissible input $u$ that will ensure $\xi$ is always within the safe set $C$; i.e., will never reach the unsafe set $\neg\mathcal{C} = \{\xi\in \mathbb{R}^n|h(\xi)<0\}$. To guarantee such forward invariance for compact sets the following minimally restrictive condition was proposed \cite{ames2017cbf},
\begin{equation}
	\exists\alpha(\cdot),\ \dot{h}(\xi) \ge -\alpha(h(\xi)) ~~ \implies \mathcal{C}~\text{is invariant},
\end{equation}
which is a necessary and sufficient condition for compact sets $\mathcal{C}$ where $\alpha(\cdot)$ is a class-$\mathcal{K}$ function that is strictly increasing and maps zero state to zero. Rewriting the above condition with the control input $u$ explicitly renders
\begin{equation}
	\nabla_\xi h(\xi)(\underline{f}(\xi) + g(\xi)u) \ge -\alpha(h(\xi)),
\end{equation}
which can be treated as an affine constraint for the following Quadratic Programming (QP) problem:
\begin{equation}
	\begin{aligned}
 \label{eq10}
		\min_{u} \quad & \|u - u_{ref}(\xi)\|_2^2\\
		\textrm{s.t.} \quad & \nabla_q h(\xi)g(\xi)u \ge -\alpha(h(\xi)) - \nabla_\xi h(\xi)\underline{f}(\xi),\\
	\end{aligned}
\end{equation}
where $u_{ref}(\xi)$ is the reference control signal given by a high-level controller. In short, this controller selects an admissible control signal that is closest to the reference signal. 

\section{PROPOSED APPROACH}
\label{section4}
Following we propose a control framework that generates control torques $\tau_c$ for a robot with dynamics \eqref{eq1} that is guaranteed to satisfy constraints $\mathcal{C}_1-\mathcal{C}_4$ defined in Section \ref{section2}, while following a desired DS motion plan $f(x)$ defined by \eqref{eq2} and allowing the robot to be passive when feasible $(\mathcal{O}_1-\mathcal{O}_2)$.
Notice that the objective of our controller is to follow a DS defined in task-space while the constraints are defined in the joint-space. Assuming we have $\mathcal{C}^2$ (at least) differentiable boundary functions that defined the invariant sets for each of the joint-space constraints; i.e., $h_{JL}^{+/-}(q):\mathbb{R}^d\rightarrow\mathbb{R}$ for joint limits, $h_{SCA}(q):\mathbb{R}^d\rightarrow\mathbb{R}$ for self-collision avoidance, $h_{ECA}(q):\mathbb{R}^d\rightarrow\mathbb{R}$ for external collision avoidance and $h_{SA}(q):\mathbb{R}^d\rightarrow\mathbb{R}$ for singularity avoidance, the general framework of the proposed QP is formulated as follows:
\begin{equation}
	\begin{aligned}
        \label{eq11}
		\min_{\tau_c} \quad & \|J(q)^{-\top}\tau_c - \underbrace{F_c(x)}_{\eqref{eq3}}\|_2^2\\
		\textrm{s.t.} \quad & H(q)\ddot{q} + C(q, \dot{q})\dot{q} + G(q) = \tau_c + \tau_{ext},\\
		\quad & \mathcal{C}_1 ~~~ \text{Joint Limit Avoidance via}~h^{+/-}_{JL}(q)\\
            \quad & \mathcal{C}_2 ~~~ \text{Self-Collision Avoidance via }~h_{SCA}(q)\\
            \quad & \mathcal{C}_3 ~~~ \text{External Collision Avoidance via}~h_{ECA}(q)\\
            \quad & \mathcal{C}_4 ~~~ \text{Singularity Avoidance via}~h_{SA}(q)\\
	\end{aligned}
\end{equation}
where $F_c$ is the DS-based passive impedance control law defined in \eqref{eq3} and  $J(q)^{-\top}$ is the pseudo-inverse of $J(q)^\top$. 

\textbf{Constraint Formulation} In the standard CBF-QP formulation \eqref{eq10} the control input $u$ has a direct relationship to the first derivative of the state $\dot{\xi}$, via \eqref{eq6}, and thus it appears in the first derivative of $h(\xi)$. In our case, the relationship between the control input $\tau_c$ and the system dynamics is with $\ddot{q}$ via \eqref{eq1}, which implies that the control signal appears in the second derivative of $h(q)$ instead of the first. This means that our constraint boundary functions $h(q)$ have relative degree of two. Thus our joint-space constraints must be formulated as exponential control barrier functions \cite{nguyen2016exponential} which require the computation of $\dot{h}_*(q)$ and $\ddot{h}_*(q)$ as will be described in Section \ref{section4a}. In Sections \ref{section4b}-\ref{section4e} we introduce the exponential CBF formulations for each of our constraints.

\textbf{QP Feasibility} It is possible for the multiple constraints above to make the QP infeasible. Therefore we first define joint limits and self-collision constraints $(\mathcal{C}_1-\mathcal{C}_2)$ as \textbf{\textit{hard constraints}} and the rest as \textbf{\textit{soft constraints}}, where hard constraints refer to constraints that are strictly prohibited to be violated and soft constraints refer to constraints that are able to be relaxed. To ensure such relaxation we formulate a relaxed version of \eqref{eq11} presented in Section \ref{section4f}.

\vspace{-5pt}
\subsection{Exponential Control Barrier Function Primer}
\label{section4a}
Considering $h(q)$ with a relative degree of two, we define a new state variable $\eta(q) = [h(q)\ \dot h(q)]^\top$. The dynamics of $\eta(q)$ can be written in the controllable canonical form:
\begin{equation}
	\begin{aligned}
	\dot \eta(q) &= A\eta(q) + B\mu,\\
	h(q) &= C\eta(q),
	\end{aligned}
\end{equation}
where $A$, $B$ and $C$ are $\left[ {\begin{array}{*{20}{c}}
		0&1\\
		0&0
\end{array}} \right]$, $[0\ 1]^\top$ and $[1\ 0]$ respectively. As proposed in \cite{nguyen2016exponential} by properly defining feedback gain vector $\mathcal{K}_f = [k_1\ k_2]$ would render forward invariance if the following inequality is satisfied:
\begin{equation}\label{eq13}
	\ddot{h}(q) \ge -\mathcal{K}_f\eta(q).
\end{equation}
Proof of forward invariance given the selection criteria of the feedback gain vector $\mathcal{K}_f$ can be referred to \cite{nguyen2016exponential}.

\subsection{Joint Limit Avoidance Constraint}
\label{section4b}
We start by deriving the constraint for joint limit avoidance. For the joint positions $q\in \mathbb{R}^n$, we expect them to stay within the limits of $q^-\leq q \leq q^+$. Hence, we define the CBFs as $h^-_{JL}(q) = q - q^- - \epsilon^-_{JL}$ for the lower and $h^+_{JL}(q) = -q + q^+ - \epsilon^+_{JL}$ for the upper limit. The corresponding QP constraints can be easily formulated as
\begin{equation}\label{eq14}
    \begin{aligned}
        &\nabla_qh^+_{JL}(q)\ddot{q} \ge -k_1h^+_{JL}(q) - k_2\nabla_q h^+_{JL}(q)\dot q \\
        &- {\dot q}^\top \nabla^2_q h^+_{JL}(q)\dot q\\
        \Rightarrow& -\ddot{q} \ge b^+_{JL}(q)\\
        & b^+_{JL}(q) = -k_1(-q + q^+ - \epsilon^+_{JL}) + k_2\dot{q}
    \end{aligned}
\end{equation}
for the upper joint limits and
\begin{equation}\label{eq15}
    \begin{aligned}
        &\nabla_qh^-_{JL}(q)\ddot{q} \ge -k_1h^-_{JL}(q) - k_2\nabla_q^- h_{JL}(q)\dot q \\
        &- {\dot q}^\top \nabla^2_q h^-_{JL}(q)\dot q\\
        \Rightarrow &\ddot{q} \ge b^-_{JL}(q) \\
         &b^-_{JL}(q) = -k_1(q - q^- - \epsilon^-_{JL}) - k_2\dot{q}
    \end{aligned}
\end{equation}
for the lower joint limits.
\vspace{-5pt}

\subsection{Self-Collision Avoidance Constraint} 
\label{section4c}
A self-collision avoidance constraint can be formulated by a continuously differentiable boundary function $\Gamma_{SCA}(q)\in \mathbb{R}^n \rightarrow \mathbb{R}$ that defines a continuous map of the safe (free) and unsafe (collided) configurations in joint-space such that:
\begin{equation}
\begin{aligned}
 \Gamma_{SCA}(q)< 0 & \implies \text{Collided configurations}\\
  \Gamma_{SCA}(q)= 0 & \implies \text{Boundary configurations}\\ \Gamma_{SCA}(q)> 0 & \implies \text{Free configurations}
\end{aligned}
\end{equation}
Explicitly defining the metric of the distance to self-collision is not straightforward. Hence, we adopt the approach proposed in prior work \cite{mirrazavi2018sca-dual,koptev2023njsdf,dsbook} that learns $\Gamma_{SCA}(q)$ by sampling the joint-space of the multi-DoF robotic systems along the boundary and uses efficient regression models (such as SVM and NN) to learn $\Gamma_{SCA}(q)$. In those works, $\Gamma_{SCA}(q)$ was only required to be $\mathcal{C}^1$ differentiable as only $\nabla_q\Gamma_{SCA}(q)$ was needed to formulate a constraint for inverse kinematics. In our work, it is also necessary that $\Gamma_{SCA}(q)$ is $\mathcal{C}^2$ differentiable in order to be exploited in an ECBF of relative degree two.  Therefore, we proposed to learn $\Gamma_{SCA}(q)$ as a Neural Network (NN) classification problem with ${\rm Tanh}$ activations (which are infinitely differentiable).

\subsubsection{Model description} Similar to \cite{koptev2021sca-humanoid} we use a Multilayer Perceptron (MLP) with 4 hidden layers, the input as robot configurations $q\in \mathbb{R}^n$, the output labeled as $[0\ 1]^\top$ for collided configurations and $[1\ 0]^\top$ for uncollided configurations. Given that the output of the NN is $[\gamma_1\ \gamma_2]^\top$ then $\Gamma_{SCA}(q) = \gamma_1 - \gamma_2$ with closed-form expression,  
\begin{equation}
\resizebox{.85\hsize}{!}{$\begin{aligned}
\begin{bmatrix}\gamma_1 \\ \gamma_2\end{bmatrix} =& \ \ \omega_5\cdot {\rm Tanh}\Big[\omega_4\cdot {\rm Tanh}\Big[\omega_3\cdot {\rm Tanh}\Big[\omega_2\\&\cdot {\rm Tanh}\Big[\omega_1\cdot q + \beta_1 \Big] + \beta_2 \Big] + \beta_3 \Big] + \beta_4 \Big] + \beta_5.
\end{aligned}$}
\end{equation}
From grid search we found a NN with the structure of 7-80-50-30-10-2 to yield the best classification accuracy for the 7DoF manipulator. The data collection strategy and learning procedure follows previous works in \cite{koptev2021sca-humanoid}, \cite{mirrazavi2018sca-dual}\footnote {\label{note2}Details on SCA boundary learning and specific constraint derivatives can be found in our project website \url{https://sites.google.com/seas.upenn.edu/constrained-passive-control}}.

\subsubsection{Constraint formulation}
With the definition of $\Gamma_{SCA}(q)$ in mind, we define the control barrier function as $h_{SCA}(q) = \Gamma_{SCA}(q) - \epsilon_{SCA}$, where $\epsilon_{SCA}$ is the minimum distance to the self-collision boundary that we expect the manipulator to maintain. Substituting the expression of $\Gamma_{SCA}$ into (\ref{eq13}) with proper $k_1$ and $k_2$:
\begin{equation}\label{eq17}
\begin{aligned}
&\ddot{h}_{SCA}(q) \ge -k_1h_{SCA}(q) - k_2\dot{h}_{SCA}(q),\\
\Rightarrow &\nabla_q\dot h_{SCA}(q)\dot q + \nabla_qh_{SCA}(q)\ddot q  \ge -k_1h_{SCA}(q)\\
&- k_2\nabla_qh_{SCA}(q)\dot q,\\
\Rightarrow &\nabla_qh_{SCA}(q)\ddot{q}  \ge b_{SCA}(q)
\end{aligned}
\end{equation}
\vspace{-5pt}
\begin{equation}
\begin{aligned}
b_{SCA}(q) &=   -k_1h_{SCA}(q) - k_2\nabla_q h_{SCA}(q)\dot q \\
&- {\dot q}^\top \nabla^2_q h_{SCA}(q)\dot q,
\end{aligned}
\end{equation}
where $\nabla_qh_{SCA}(q) \in \mathbb{R}^{n}$ and  $\nabla^2_qh_{SCA}(q)\in \mathbb{R}^{n\times n}$ denote the gradient vector and Hessian matrix of $h_{SCA}(q)$\footref{note2}.
\vspace{-5pt}
\subsection{External Object Collision Avoidance Constraint}
\label{section4d}
Adopting the work of \cite{koptev2023njsdf} we utilize their Neural Joint-Space Signed Distance Function (Neural-JSDF) for external object collision avoidance to formulate a CBF constraint. 
\subsubsection{Brief Introduction to JSDF}
SDFs represent the surface of objects and the minimal distance from a query point to the surface through a continuous second-derivable function. We define the SDF of a robot manipulator in joint-space $\Gamma_{SDF}(q, x_0)\in \mathbb{R}^n_L$ as $\Gamma_{SDF}^i(q, x_0) = \min\|x - x_0\|_2,\ i = 1\cdots n_L$, where $x_0\in \mathbb{R}^3$ is the query point in the task-space, $x$ represent points on the surface: $\{x\in \mathbb{R}^3|\Gamma_{SDF}(q, x) = 0\}$ in task-space, $n_L$ denotes the number of links. $\Gamma_{SDF}^i(q, x_0)$ is the $i$-th element of $\Gamma_{SDF}^i(q, x_0)$ which denotes the minimal distance corresponding to the $i$-th link. Due to the complexity of the surface geometry of the robot manipulators, a data-driven approach, MLP, is used to fit $\Gamma_{SDF}(q, x_0)$.  We use the learned Neural-JSDF model for the Franka arm provided in the accompanying code of \cite{koptev2023njsdf}, refer to paper for model description, data generation, and training procedures\footnote{\url{https://github.com/epfl-lasa/Neural-JSDF}}.

\subsubsection{Constraint formulation}
Similar to the ECBF for self-collision avoidance, the ECBF for external object collision avoidance is defined as $h^{(i)}_{ECA}(q) = \Gamma^{(i)}_{SDF}(q, x_0) - \epsilon_{SCF}$ for the $i$-th link and the external query point as $x$. Consequently, the QP with the constraints of external object collision avoidance is formulated for every link $i=1\dots,n_L$ as:
\begin{equation}\label{eq19}
	\begin{aligned}
		\quad & \nabla_qh^{(i)}_{ECA}(q)\ddot{q} \ge b_{ECA}^{(i)}(q)
	\end{aligned}
\end{equation}
\begin{equation}\label{eq19}
	\begin{aligned}
		b_{ECA}^{(i)}(q) = &  -k_1h^{(i)}_{ECA}(q) - k_2\nabla_q h^{(i)}_{ECA}(q)\dot q \\
        &- {\dot q}^\top \nabla^2_q h^{(i)}_{ECA}(q)\dot q.\\
	\end{aligned}
\end{equation}

\subsection{Singularity Avoidance}
\label{section4e}
The ability of a robot to stay far from joint singularities is directly related to its manipulability \cite{Yoshikawa1985Manipulability}. Among the several metrics that can be used to measure manipulability, like the weighted Frobenius norm of $J(q)$ and the manipulability ellipsoid, in this work we choose the manipulability index.

\subsubsection{Brief Introduction to Manipulability Index}
The manipulability index $\mathcal{MI}(q) \in \mathbb{R}^n \rightarrow \mathbb{R}$ is computed as 
\begin{equation}\label{eq4}
    \mathcal{MI}(q) = \sqrt{\det[J(q)J^\top(q)]} = \prod_{i = 1}^{d}\sigma_i
\end{equation}
where $\sigma_1, \dots, \sigma_d$ are the singular values of $J$ extracted from the singular value decomposition of $J(q)$; i.e.,  $J = U\Sigma V^\top$ \cite{spong2nd}. This index is well suited to formulate a constraint regarding singularity avoidance because when it increases,
the manipulability of the robot does as well, and when it
gets close to 0, the robot is close to a singularity.
\subsubsection{Constraint formulation}
To keep the manipulability index greater than its lower bound $\epsilon_{SA} > 0$, we define the CBF for singularity avoidance $h_{SA}(q)$ as $h_{SA}(q) = \mathcal{MI}(q) - \epsilon_{SA}$, and the constraint is formulated as:
\begin{equation}\label{eq7}
\nabla_qh_{SA}(q)\ddot{q} \ge  b_{SA}(q)
\end{equation}
\begin{equation}
\begin{aligned}
    b_{SA}(q) = & -k_1h_{SA}(q) - k_2\nabla_q h_{SA}(q)\dot q \\
    & - {\dot q}^\top \nabla^2_q h_{SA}(q)\dot q
\end{aligned}
\end{equation}
The gradient of $\mathcal{MI}(q)$ w.r.t the $i$-th element ($i = 1\cdots n$) of $q$ has the following closed-form expression:
\begin{equation}
    \nabla_{q_i} \mathcal{MI}(q) = \mathcal{MI}(q){\rm \mathbf{Tr}}\big\{{(JJ^\top)^{-1}(\nabla_{q_i}J)J^\top}\big\},
\end{equation}
where ${\rm \mathbf{Tr}\{\cdot\}}$ denotes the trace of a given matrix\footref{note2}. Nevertheless, deriving the closed-form expression of the Hessian of $\mathcal{MI}(q)$ is more complicated. An alternative is to compute $\nabla^2_q\mathcal{MI}(q)$ via numerical approximation:
\begin{equation}\label{eq22}
    \resizebox{.89\hsize}{!}{$\nabla^2_{q_i}\mathcal{MI}(q) \buildrel\textstyle.\over = \frac{\nabla\mathcal{MI}(q + \Delta q_i e_i) - \nabla\mathcal{MI}(q - \Delta q_i e_i)}{\Delta q_i},$}
\end{equation}
where $\nabla^2_{q_i}\mathcal{MI}(q)$ is the second-order gradient w.r.t. the $i$-th element of $q$, $e_i \in \mathbb{R}^n$ is a indicator vector with its $i$-th element as 1 and other elements as 0,  $\Delta q_i \in \mathbb{R} > 0$ is a small step of $q_i$. Note that in real applications, in order to compute the parameters in real-time, employing (\ref{eq22}) is not practical due to its computation complexity. To address this issue, we re-write the term ${\dot q}^\top \nabla_q^2h_{SA}(q)\dot q$ in (\ref{eq7}) as $\dot{\nabla_q h_{SA}}(q)\dot q$, where 
\begin{equation}
    \dot{\nabla_q h_{SA}}(q) \buildrel\textstyle.\over =  \frac{\nabla \mathcal{MI}(q_t) -  \nabla\mathcal{MI}(q_{t-1})}{\Delta t}.
\end{equation}

\subsection{Passive Interaction Control with Constraints}
\label{section4f}
Now that we have introduced the formulation of all constraints $\mathcal{C}_1-\mathcal{C}_4$ as ECBFs we can re-state our QP optimization problem \eqref{eq11} as a Relaxed-CBF-QP (R-CBF-QP) \cite{lee2023hierarchical} in order to ensure prioritization of \textbf{hard constraints}:
\begin{equation}
\label{eq:final_qp}
	\begin{aligned}
		\min_{\tau_c, \delta_{ECA}, \delta_{SA}} \quad & \|J(q)^{-\top}\tau_c - \underbrace{F_c(x)}_{\eqref{eq3}}\|_2^2 + \delta_{ECA}^\top \Pi \delta_{ECA} + \pi \delta^2_{SA}\\
		\textrm{s.t.} \quad & H(q)\ddot{q} + C(q, \dot{q})\dot{q} + G(q) = \tau_c + \tau_{ext},\\
        \mathbf{\mathcal{C}_1}: &~~ -\ddot q \ge b^+_{JL}(q), ~~ \ddot q \ge b^-_{JL}(q)\\
	\mathbf{\mathcal{C}_2}: &~~  \nabla_q h_{SCA}(q) \ddot q \ge b_{SCA}(q)\\
        \mathcal{C}_3:& ~~ \nabla_q h^{(i)}_{ECA}(q) \ddot q \ge b_{ECA}^{(i)}(q) - \delta_{ECA}^{(i)}\\
        \quad &  ~~~ \delta_{ECA}^{(i)} \ge 0 ~~\forall i=1,\dots, n_L\\
        \mathcal{C}_4:&~~  \nabla_q h_{SA}(q) \ddot q \ge b_{SA}(q) - \delta_{SA}\\
        \quad &~~ \delta_{SA} \ge 0,
	\end{aligned}
\end{equation}
where $\delta_{ECA} = [\delta_{ECA}^{(1)}, \cdots, \delta_{ECA}^{(n_L)}]^\top \in \mathbb{R}^{n_L}$ and $\delta_{SA}$ are the slack variables for \textit{soft constraints}. $\Pi \succ 0$ and $\pi>0$ relax the constraints that are defined by preference.

\begin{rem}[Passivity]
A robot with system dynamics \eqref{eq1} under control input $\tau_c$ optimized by \eqref{eq:final_qp} and following a task-space control input $F_c$ by \eqref{eq3} is passive if an optimal $\tau_c^*$ can be found in the feasible set, i.e., when $J(q)^{-\top}\tau_c =F_c(x)$ or $F_c \notin \mathcal{N}(J^{-\top})$ -- \textit{passive when feasible}.
\end{rem}

\begin{figure*}[!tbp]
\centering
\includegraphics[trim={8.5cm 0cm 10cm 0cm},clip,width=0.315\textwidth]{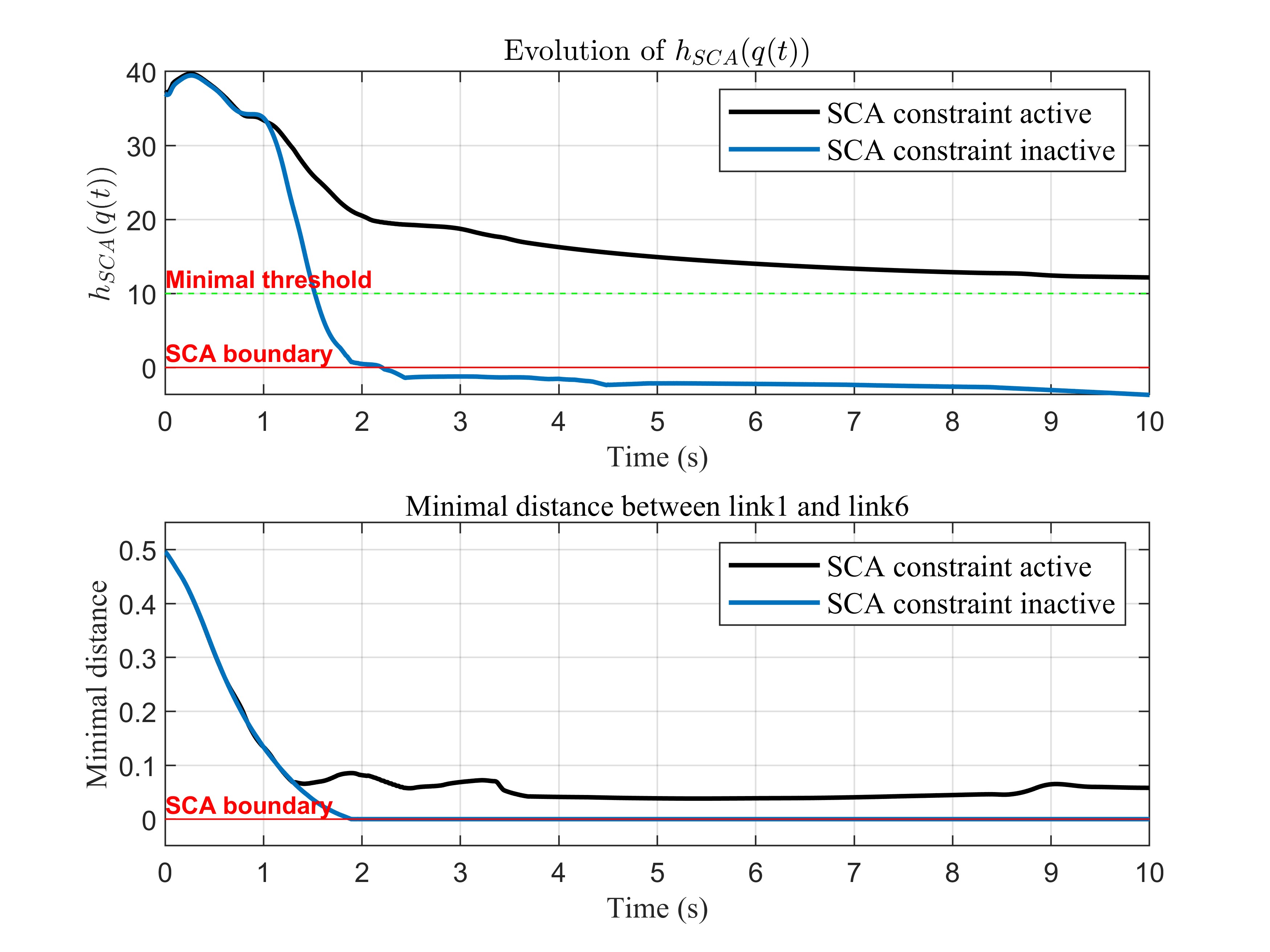}
\includegraphics[trim={6.5cm 0cm 10cm 0cm},clip,width=0.315\textwidth]{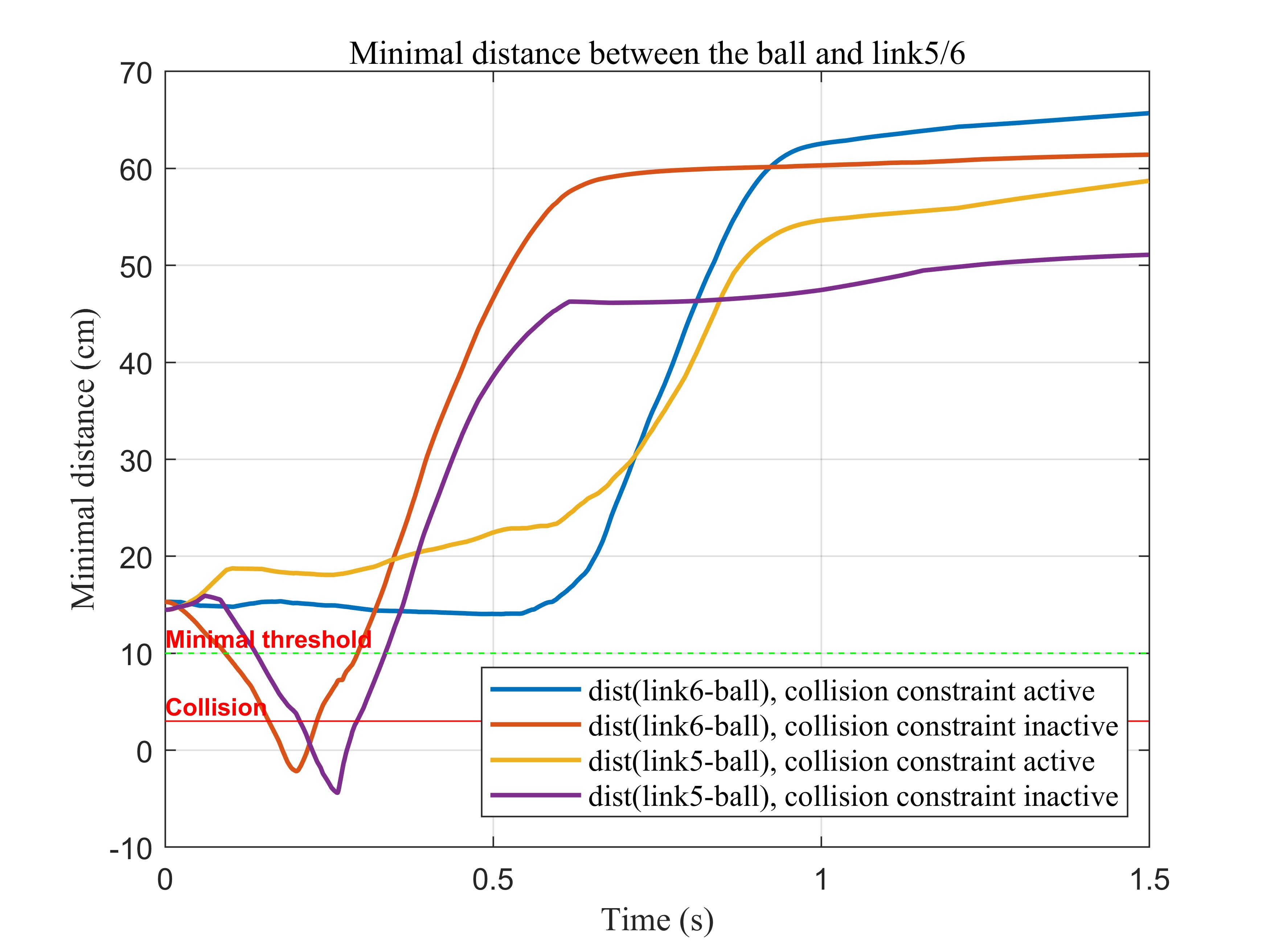}
\includegraphics[trim={4.5cm 0cm 11cm 0cm},clip,width=0.315\textwidth]{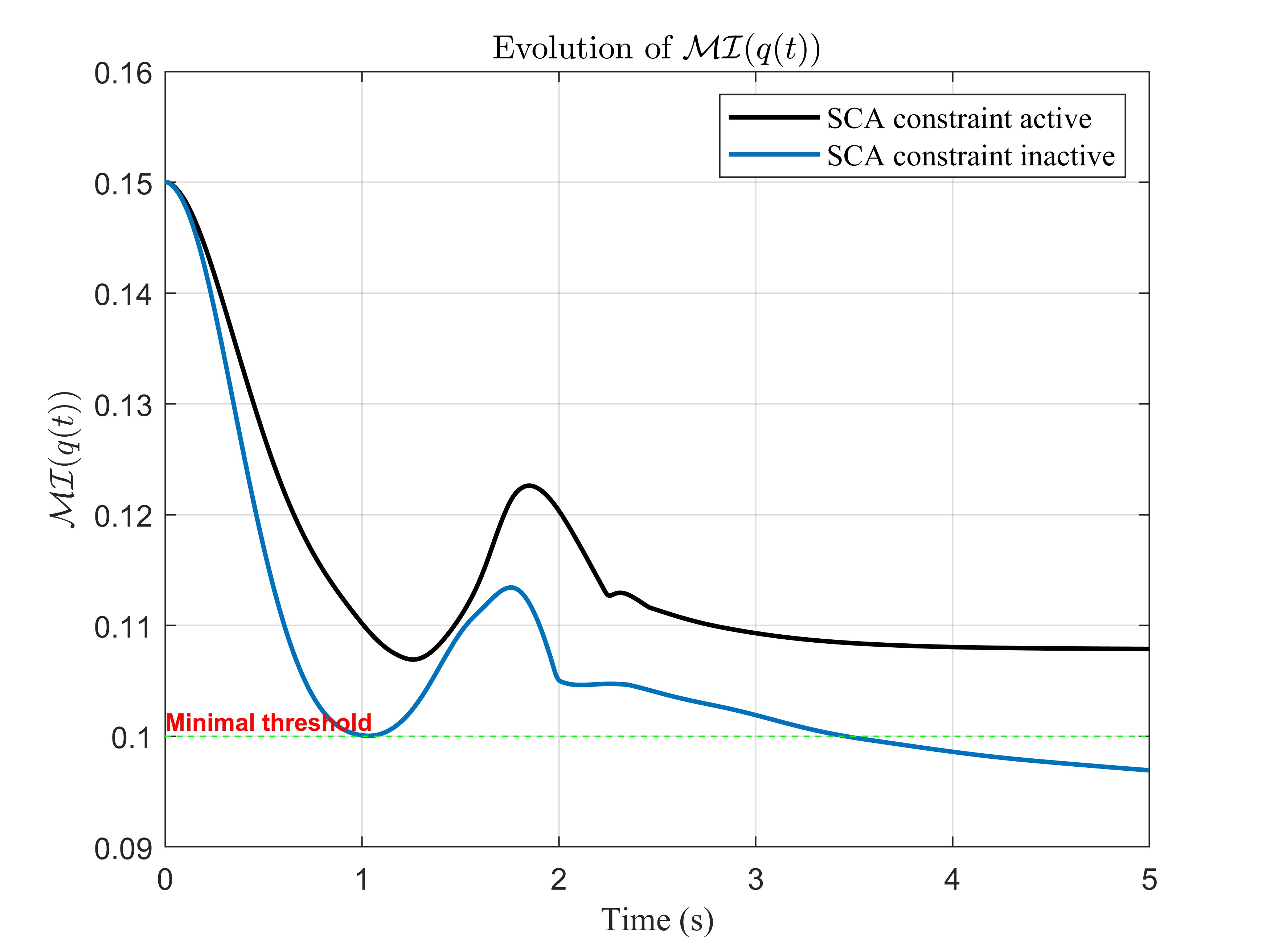}
\vspace{-10pt}
\caption{\small Evolution of joint-space boundary functions with active and inactive constraints on the 7-DoF Franka Research 3 manipulator for the simulated tests in Section \ref{section5a}. (left) Evolution $h_{SCA}$ and the minimal distance between link1 and link6 of a. (center) Minimal distance between the robot and the external ball with and without the external object collision constraints. Only link6 and link5 are reported, which tend to collide with the ball that we defined without the constraints. (right) Evolution of Manipulability Index $\mathcal{MI}(q)$.  \label{fig:simulation_plots}}
\vspace{-10pt}
\end{figure*}
\begin{figure*}[!tbp]
\includegraphics[trim={7.5cm 0cm 10cm 0cm},clip,width=0.31\textwidth]{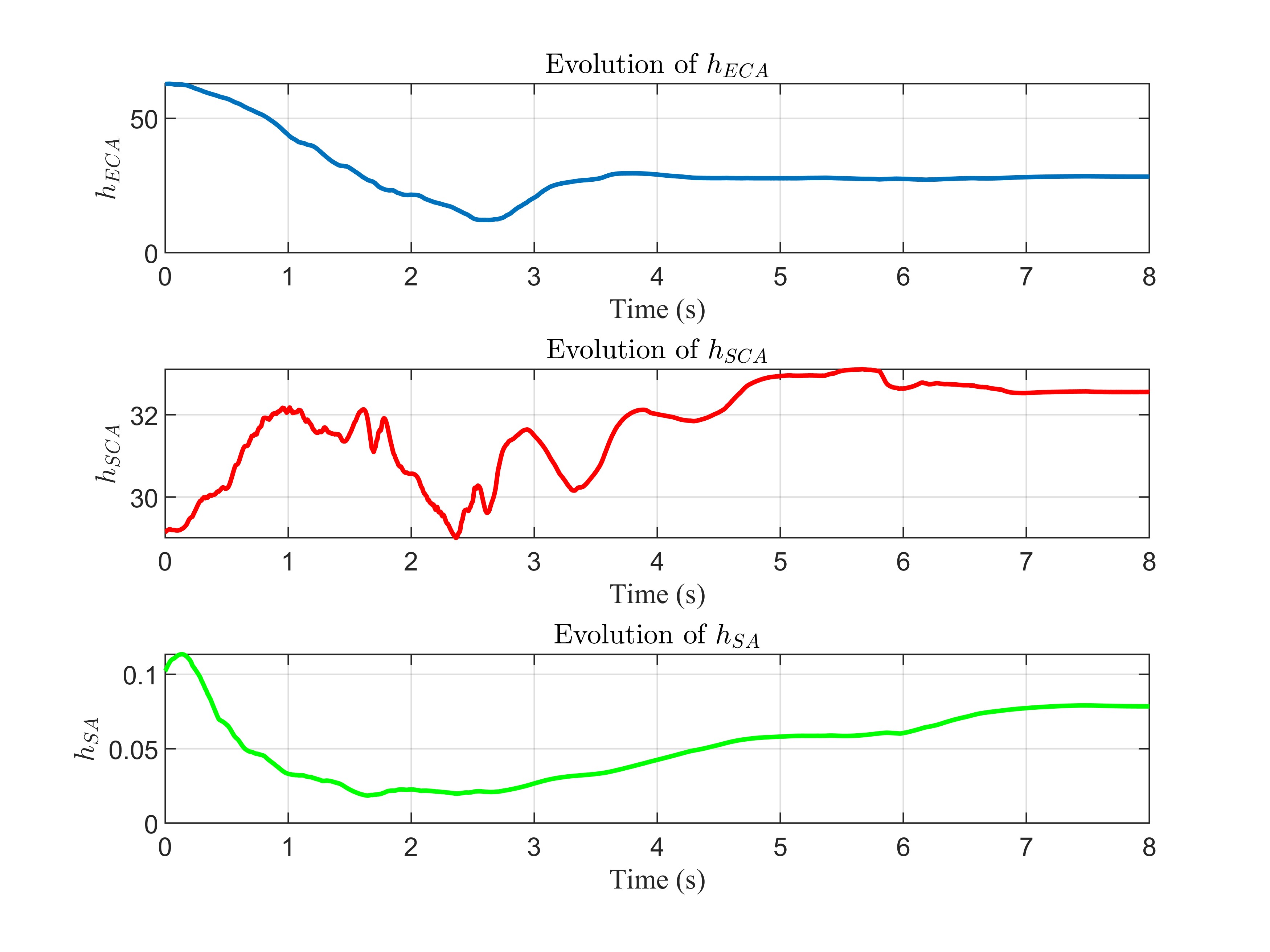}
\includegraphics[trim={0cm 0cm 0cm 0cm},clip,width=0.35\textwidth]{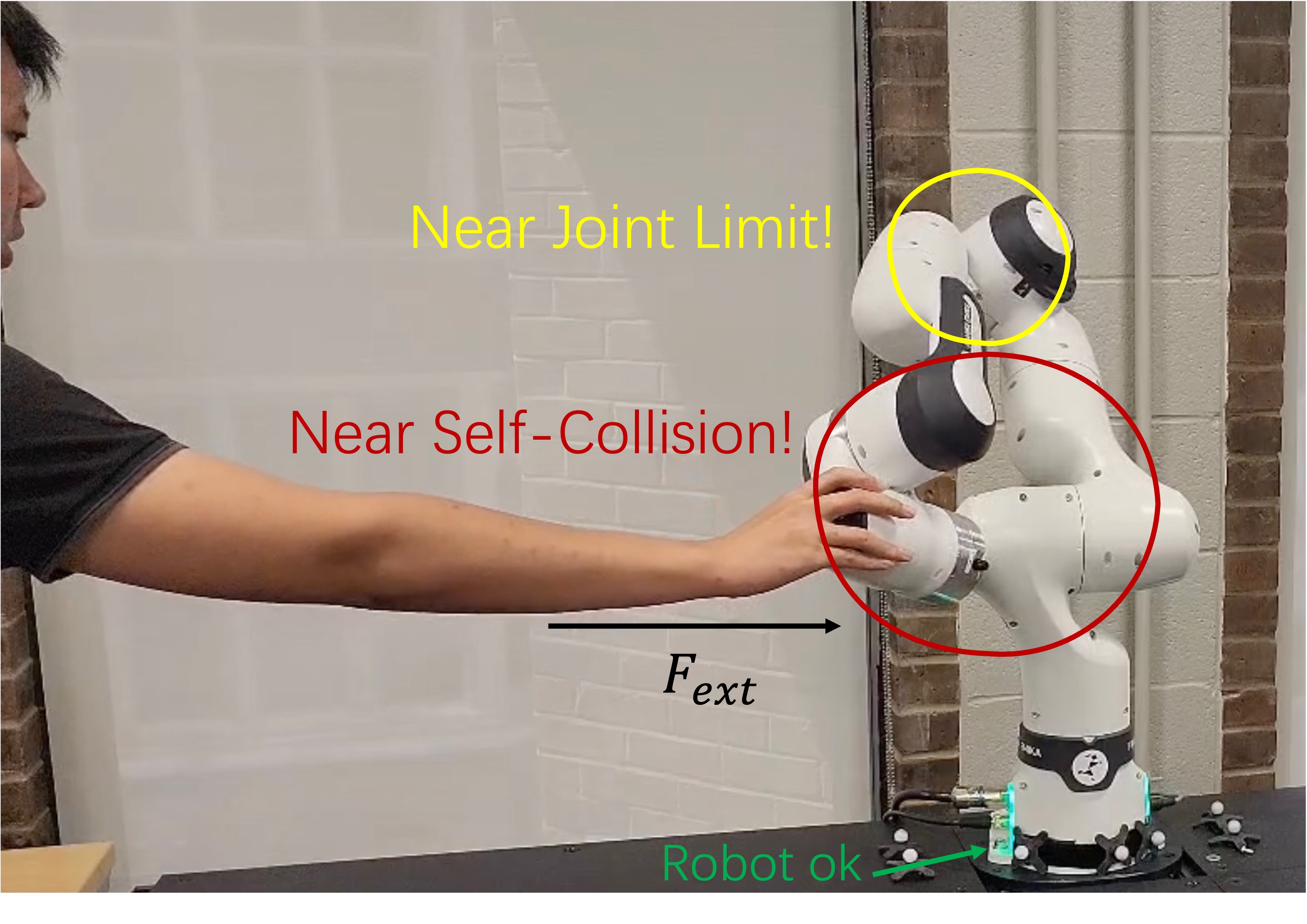}
\includegraphics[trim={6.5cm 0cm 10cm 0cm},clip,width=0.31\textwidth]{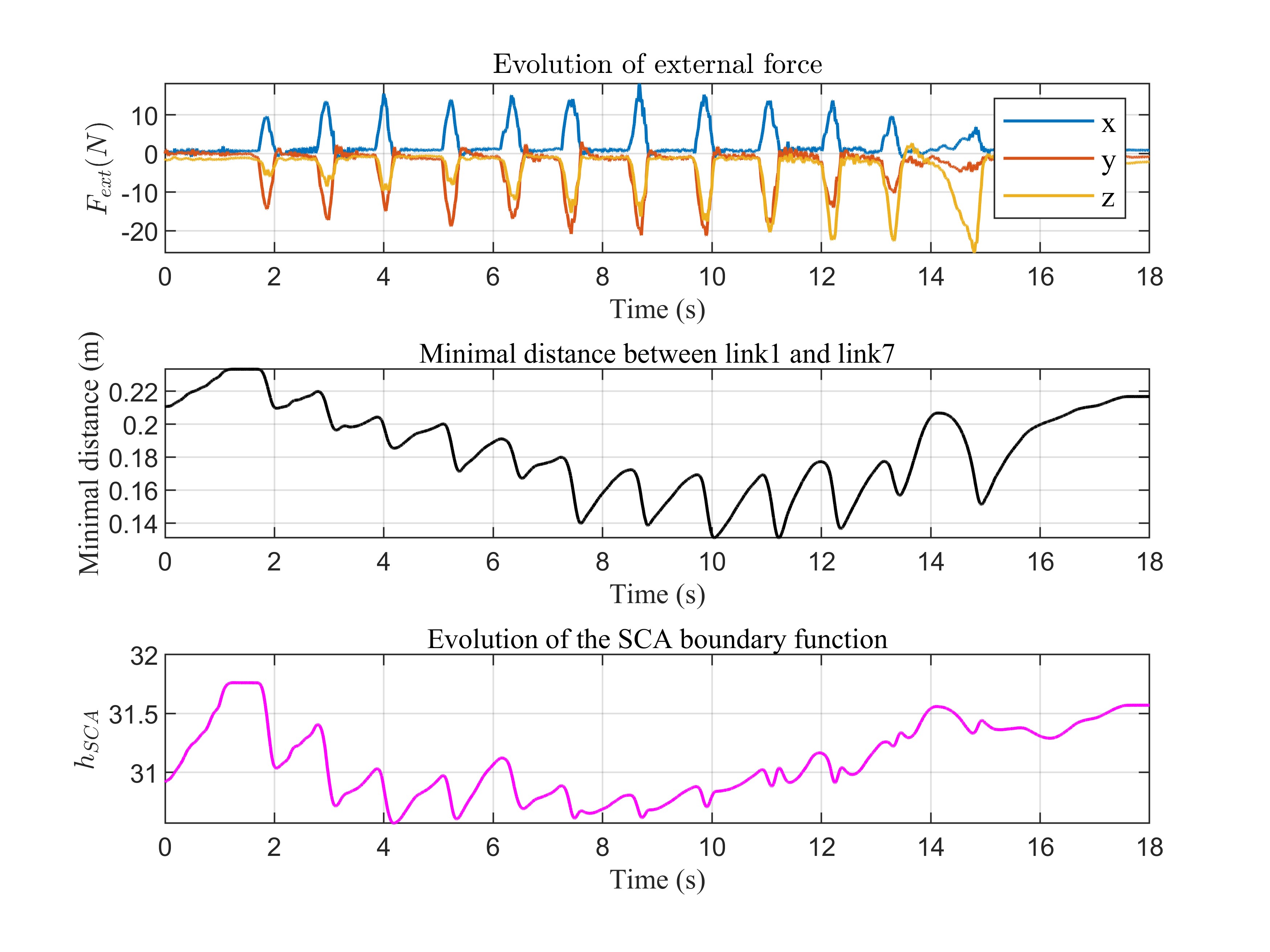}
\vspace{-8pt}
\caption{\small (left) Evolution of all joint-space boundary functions for the simulated tests in Section \ref{section5a}. (center) Controller on real-robot allowing external perturbations yet satisfying self-collision and joint limit constraints. (right) $F_{ext}$ and $h_{SCA}$ for real-robot experiment. \label{fig:more_plots}}
\vspace{-10pt}
\end{figure*}

\section{EXPERIMENTAL RESULTS}
\label{section5}
Videos of PyBullet simulation and hardware experiments are included in the accompanying multimedia submission and can be found in our project website\footref{note2}. All experiments are conducted on a 7-Dof Franka Research 3 manipulator. The QP optimization solver is implemented in CVXPY \cite{diamond2016cvxpy} for simulations and CVXGEN \cite{MattingleySB12} for real robot experiments.
\vspace{-2pt}
\subsection{Simulation Experiments}
\label{section5a}
We design four experiments to evaluate the satisfaction of combinations of 1) self-collision avoidance + joint limits, 2) collision avoidance to external objects + joint limits, 3) singularity avoidance + joint limits using CBF-QP, and 4) all constraints and adopt the R-CBF-QP.  

\subsubsection{Self-Collision Avoidance \& Joint Limits}
In this scenario, we seek to find a DS $f(x)$ with convergence point inside the body of the manipulator. Obviously, the robot will collide with itself if the self-collision constraint is not added. We therefore simply define the potential function as:
\begin{equation}
    V(x) = (x - x^*)^\top P (x - x^*),
\end{equation}
where $x^* = [0, 0, 0.3]^\top$ and $P = -25\mathbb{I}_3$, and $f(x)$ as:
\begin{equation}
    \dot x = f(x) = \nabla_xV(x) = 2P(x - x^*)
\end{equation}
which is linear and conservative. In this testing scenario, we define a threshold of 10 that we expect the manipulator to keep above to. Fig. \ref{fig:simulation_plots} (left) shows that the manipulator reaches to self-collision configurations without the SCA avoidance constraint and link1 and link6 collide with each other. By introducing the self-collision constraint, link1 and link6 keep a minimum distance from each other implicitly via $h_{SCA}(q)$. The simulation lasts 10 seconds.

\subsubsection{Collision Avoidance to External objects \& Joint Limits}
We consider a simple scenario where there is a static ball as an obstacle on the route of the robot to its convergence point $x^* = [-0.25,\ -0.35,\ 0.5]^\top$, the position of the ball is $[0.35,\ -0.15,\ 0.7]^\top$.The simulation lasts 1.5 seconds. The manipulator is expected to keep at least 10 cm away from the obstacle. Fig. \ref{fig:simulation_plots} (center) shows that collisions to external objects are guaranteed with the corresponding constraints. 

Further, in the accompanying video we showcase a scenario where the manipulator is exposed to external perturbations. The DS pulls the robot back to its initial point $x^* = [0.7, 0, 0.5]^\top$, and the ball with its radius of 10 cm is set at $[0.25, -0.25, 0.6]^\top$. An external force of $F_{ext} = [0, -10, 0]^\top$ is added to the end-effector on the first 0.8 secs. The controller avoids collision when external force exists.

\subsubsection{Singularity Avoidance \& Joint Limits}
In order to verify the effect of the singularity avoidance constraints that guarantee the robot manipulability, we designed a situation where the manipulability index reduces below 0.1, which is our minimal threshold. $x^* = [-0.6,\ 0,\ 0.4]^\top$. Consequently, the controller with the constraint of singularity avoidance maintains the manipulability index to approximately 0.11, as shown in Fig. \ref{fig:simulation_plots} (right). The simulation lasts 5 seconds.

\subsubsection{ALL Constraints} We design a scenario where the robot enters a self-collision configuration whilst entering a `dangerous' zone which is close to a sphere at $[0.25, -0.25, 0.3]^\top$ and show that th robot controller drives the robot to converge at $x^*[0, -0.5, 0.3]^\top$, while satisfying all constraints as shown in Fig. \ref{fig:more_plots} (left). For the sake of reducing computation consumption's, only link7 is considered for $h_{ECA}(q)$.

\subsection{Real-Robot Experiments}
In a real-robot experiment, we showcase both SCA and joint-limits constraint satisfaction while being passive and subject to external perturbations. The manipulator is guided by linear DS stabilized at $x^*=[0, -0.5, 0.3]^\top$. Then external force is added on the end-effector of the robot. Fig. \ref{fig:more_plots} shows that under human perturbation, the passive controller still has the performance of self-collision avoidance. The computation frequency related to NNs and QP are approx. 300Hz and 200Hz. The controller operates at 1kHz.

\section{CONCLUSIONS AND FUTURE WORKS}\label{section6}
We proposed a novel framework for constrained passive interaction control of robot manipulators. We leverage safety and passivity by introducing joint-space constraints on a task-space DS-based passive controller via an exponential CBF approach. We showcased the effectiveness of the controller through simulation and real-robot experiments. Currently our computational bottleneck is the computation of the Hessian for the NN-based JSDF functions. Hence, we plan to explore more model efficient boundary functions such as \cite{li2023learning} as well as evaluate our controller on practical applications.




\newpage
\bibliographystyle{IEEEtran}
\bibliography{ref.bib}

\end{document}